%% file: arxiv_deblurSceneFlow.tex
\documentclass[runningheads]{llncs}
\usepackage{graphicx}
\usepackage{amsmath,amssymb} 
\usepackage{color}
\usepackage[width=122mm,left=12mm,paperwidth=146mm,height=193mm,top=12mm,paperheight=217mm]{geometry}

\usepackage[format=plain,labelformat=simple,labelfont=bf,labelsep=period,font=small,skip=4pt,compatibility=false]{caption}
\usepackage[labelfont=rm,font=scriptsize]{subfig}
\usepackage{booktabs}
\usepackage{tabu}
\usepackage{mathtools}
\usepackage{array}
\usepackage{nicefrac,siunitx}
\usepackage{times}
\usepackage[super]{nth}

\usepackage{xspace}
\newcommand{\eg}{\emph{e.\thinspace{}g.}\@\xspace}
\newcommand{\ie}{\emph{i.\thinspace{}e.}\@\xspace}

\newcommand{\etal}{\emph{et al.}\@\xspace}

\newcommand{\Eq}{Eq.\@\xspace}

\begin{document}
\pagestyle{headings}
\mainmatter
\def\ECCV16SubNumber{403}  

\title{Stereo Video Deblurring} 

\titlerunning{Stereo Video Deblurring}

\authorrunning{Sellent, Rother, and Roth}

\author{Anita Sellent\inst{1,2}  \and Carsten Rother\inst{1} \and Stefan Roth\inst{2}}
\institute{Technische Universit\"at Dresden, Germany \and Technische Universit\"at Darmstadt, Germany }

\maketitle

\begin{abstract}
\input{abstract}
\keywords{object motion blur, scene flow, spatially-variant deblurring}
\end{abstract}

\input{introduction}

\input{related}

\input{technical}

\input{experiments}

\input{conclusion}

\clearpage
\bibliographystyle{splncs}
\bibliography{short,papers,external,motionBlurStereo}

\input{arxiv_supplement}
\end{document}

%% file: abstract.tex
Videos acquired in low-light conditions often exhibit motion blur, which
depends on the motion of the objects relative to the camera.
This is not only visually unpleasing, but can hamper further
processing. 
With this paper we are the first to 
show how the availability of stereo video can aid the
challenging video deblurring task.
We leverage 3D scene flow, which can be estimated robustly even under
adverse conditions. 
We go beyond simply determining the object motion in two ways:
First, we show how a piecewise rigid 3D scene flow representation
allows to induce accurate blur kernels via
local homographies.
Second, we exploit the estimated motion boundaries of the 3D scene flow 
to mitigate ringing artifacts
using an iterative weighting scheme. 
Being aware of 3D object motion, our approach can deal robustly with
an arbitrary number of independently moving objects.
We demonstrate its benefit over state-of-the-art video deblurring
using quantitative and qualitative experiments on rendered scenes and
real videos.

%% file: introduction.tex
\section{Introduction}
Stereo is one of the oldest areas of computer vision research
\cite{Longuet:1981:CAR}.
Interestingly, the arrival of mass-produced active depth sensors
\cite{Shotton:2013:EHP} seems to have renewed interest also in passive
stereo systems.
In contrast to active depth sensors, stereo cameras are also
applicable in outdoor environments.
Due to their more general applicability,
stereo cameras are gaining increased adoption, for example in autonomous
driving~\cite{Franke:2000:RTS}.
Remarkably, the availability of stereo image pairs also helps in the
estimation of temporal correspondences:
On the KITTI optical flow benchmark \cite{Geiger:2012:AWR}, the best
performing algorithms \cite{Menze:2015:OSF,Vogel:2015:3SF} are indeed
scene flow algorithms that jointly estimate depth and 3D motion from
stereo videos.
Part of their advantage stems from an increased robustness to adverse
imaging conditions \cite{Vogel:2015:3SF}.
One such adverse imaging condition is a shortage of light.
In low-light conditions, the exposure time often needs to be
increased to obtain a reasonable signal-to-noise-ratio.
But when either the camera, or the objects in the scenes
are moving during exposure time, this results in motion blurred images.
\input{fig_teaser}
\input{fig_sceneFlow}

Motion blur is not only unsatisfactory to look at, it can also disturb
further image-based processing, \eg in tasks such as
panorama stitching \cite{li2010generating} or barcode recognition
\cite{yahyanejad2010removing}.
In stereo video setups, viewpoint-dependent motion blur hinders a post-capture adjustment of the baseline, the acquisition and visualization of
3D point clouds (see Fig.~\ref{fig:teaser} for an example) or the controll of tele-operated robots in the presence of rapid robot and/or object motion.

%

%
In this paper we address the challenge of deblurring stereo videos.
In contrast to the substantial literature on removing camera shake
\cite{Fergus:2006:RCS,Cho:2009:FMD,Whyte:2010:NDS,Krishnan:2011:BDN,Xu:2013:ULS,Michaeli:2014:BDI},
we aim to deal with the more general case of camera \emph{and} object
motion.
In case of independent motions, mixed pixels at motion boundaries yield significant complications.
Removing such spatially-variant blur is extremely challenging when
attempted from single images
\cite{Schelten:2014:LIB,Couzinie-Devy:2013:LER}, but video input helps
to significantly increase robustness
\cite{Kim:2015:GVD,Wulff:2014:MBV}.
Unlike previous work, we leverage stereo video to obtain substantially
improved and more robust deblurring results.
In our approach, we exploit 3D scene flow in various ways and \emph{make the
following contributions:}
\emph{(i)} We show that 3D scene flow can improve video deblurring by
providing more accurate motion estimates.
In particular, we exploit piecewise rigid scene flow
\cite{Vogel:2015:3SF}, which yields an over-segmentation of the image
into planar patches that move with a rigid 3D motion
(Figs. \ref{fig:sceneFlow:overseg} and \ref{fig:sceneFlow:OF}).
\emph{(ii)}
We demonstrate that the resulting piecewise homographies
allow to directly induce blur matrices.
Thereby, we take into account that the projection of a rigid 3D
motion yields non-linear motion trajectories in 2D (Fig.~\ref{fig:kernel}, Tab.~\ref{tab:kernel}).
We find that this leads to superior deblurring results compared to
inducing the blur matrices from an optical flow field \cite{Kim:2015:GVD}
(Figs.\ref{fig:sceneFlow:2DSF} to \ref{fig:sceneFlow:Kim}).
\emph{(iii)} We apply the homography-induced blur matrices in
a robust deblurring procedure that attenuates the effects of motion
discontinuities using an iterative weighting scheme;
Initial motion discontinuities are obtained from 3D scene flow.
We demonstrate the superiority of the proposed stereo video deblurring
over state-of-the-art monocular video deblurring using experiments on
synthetic data as well as on real videos.


%% file: fig_teaser.tex
\begin{figure}[t]
\centering

\subfloat[Input]{\label{fig:teaser:input}\includegraphics[width = 0.20\columnwidth]{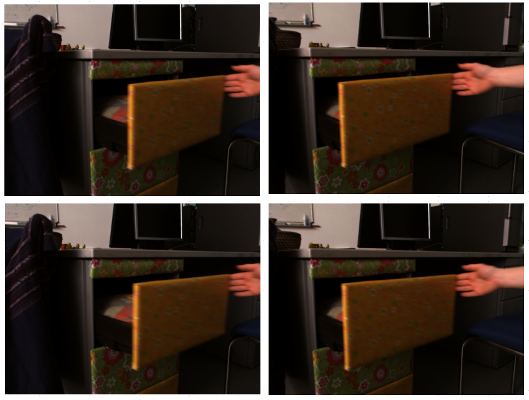} }%
\subfloat[Blurry 3D point cloud]{\label{teaser:blurry}\includegraphics[width = 0.40\columnwidth]{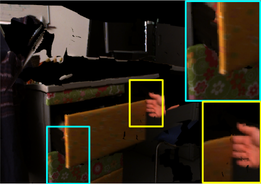} }%
\subfloat[Deblurred 3D point cloud]{\label{teaser:sharp}\includegraphics[width = 0.40\columnwidth]{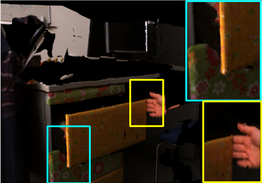} }%
\caption{Application of stereo video deblurring: Given 2 consecutive stereo frames~\protect\subref{fig:teaser:input}, our deblurring approach allows to estimate sharp textures 
  from stereo video input with motion blur. Rendering scene flow geometry with the blurred input image as a colored point-cloud from a new point of view produces an unnatural motion blur~\protect\subref{teaser:blurry}. Our stereo video deblurring algorithm can remove the blur~\protect\subref{teaser:sharp}}
\label{fig:teaser}
\end{figure}

%% file: fig_sceneFlow.tex
\begin{figure}[t]
\centering
\subfloat[Input: 2 stereo frames]{\label{fig:sceneFlow:input}\includegraphics[width = 0.32\columnwidth]{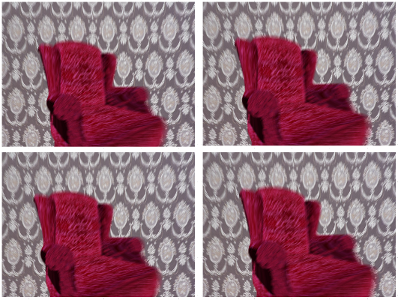} }%
\subfloat[Over-segmentation]{\label{fig:sceneFlow:overseg}\includegraphics[width = 0.32\columnwidth]{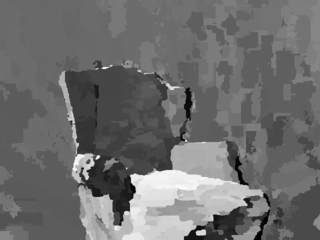} }%
\subfloat[Optical flow]{\label{fig:sceneFlow:OF}\includegraphics[width = 0.32\columnwidth]{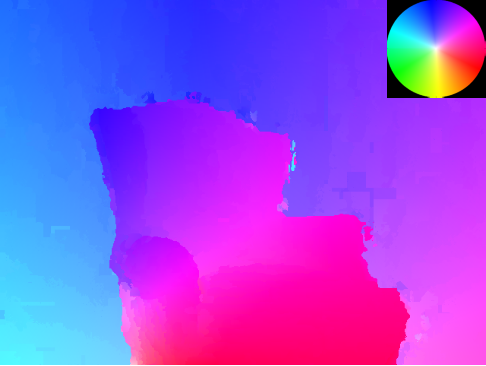} }%
\\
\vspace{-0.35cm}
\subfloat[Optical flow deblurring]{\label{fig:sceneFlow:2DSF}\includegraphics[width = 0.32\columnwidth]{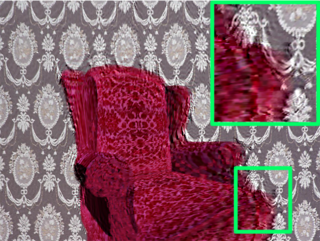} }%
\subfloat[Ours]{\label{fig:sceneFlow:3DSF}\includegraphics[width = 0.32\columnwidth]{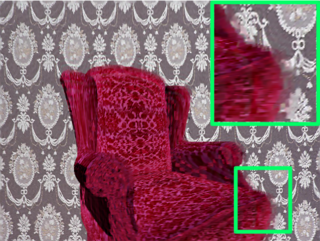} }%
\subfloat[Video deblurring~\cite{Kim:2015:GVD}]{\label{fig:sceneFlow:Kim}\includegraphics[width = 0.32\columnwidth]{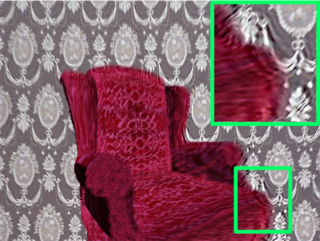} }%
\caption{Stereo video deblurring: For two consecutive frames of a synthetic stereo
  video~\protect\subref{fig:sceneFlow:input} we use the scene flow  
  approach of Vogel \etal~\cite{Vogel:2015:3SF} to compute an
  over-segmentation into planar patches with constant 3D rigid body
  motion~\protect\subref{fig:sceneFlow:overseg}.
  Projecting the 3D motion onto the image plane yields
  optical flow~\protect\subref{fig:sceneFlow:OF}, which our baseline algorithm uses to
  deblur a reference frame~\protect\subref{fig:sceneFlow:2DSF}.
  Exploiting the homographies from the 3D motion and object boundary
  information from the over-segmentation, our full approach obtains sharp
  images avoiding ringing and boundary
  artifacts~\protect\subref{fig:sceneFlow:3DSF}.
  Our result is also clearly sharper than state-of-the-art video
  deblurring \cite{Kim:2015:GVD}~\protect\subref{fig:sceneFlow:Kim}}
\label{fig:sceneFlow}
\end{figure}

%% file: related.tex
\section{Related Work}
\input{fig_kernel}
The goal of this work is to obtain sharp images from stereo videos
containing 3D camera and object motion. 
Of course, in principle blind deblurring could be applied to each frame
individually. 
However, blind motion deblurring from a single image is a
highly underconstrained problem, as blur parameters and sharp image have
to be estimated from a single measurement. 
To cope with spatially-variant blur due to the 3D motion of
the camera, single image deblurring approaches frequently use
homographies~\cite{tai2011richardson,gupta2010single,rajagopalan2014motion}.
In contrast, we apply homographies to describe spatially-variant
object motion blur.
Single image object motion deblurring approaches keep the number of
parameters manageable by either choosing the motion of a region from a
very restricted set of spatially-invariant box filters
\cite{Levin:2006:BMD,Chakrabarti:2010:ASB}, assuming it to have a
spatially-invariant, non-parametric kernel of limited size
\cite{Schelten:2014:LIB}, or to be representable by a discrete set of
basis kernels \cite{Kim:2013:DSD}.
Approaches that rely on learning spatially-variant blur are also
limited to a discretized set of detectable motions
\cite{Couzinie-Devy:2013:LER,Sun:2015:LCN}.
Kim \etal \cite{Kim:2014:SFD} consider continuously varying box filters for every pixel,
but rely heavily on regularization.
Connecting deblurring and depth estimation, 
Xu and Jia \cite{Xu:2012:DAM} successfully apply stereo correspondence estimation to
motion-blurred stereo frames to support blind image deblurring.
Lee and Lee \cite{lee2013dense}, Arun
\etal~\cite{arun2015multi}, and Hu \etal~\cite{hu2014joint} estimate sharp images and depth jointly.
However, all these approaches assume the scene to be static and camera
motion to be the only source of motion blur.

Cho \etal~\cite{cho2007removing} deblur images of
independently moving objects.
The multiple input images of their algorithm are unordered, and a
piecewise affine registration between the images, as well as the motion
underlying the blur, has to be estimated.
To restrict the parameter space, the blur kernels are assumed to be
piecewise constant and linear.

Video deblurring approaches reduce the number of parameters through
the assumption that the inter-frame and intra-frame motion are
related by the duty cycle of the camera.
He \etal~\cite{he2010motion} and Deng \etal~\cite{deng2012video} apply
feature tracking of a single moving object to obtain
2D displacement-based blur kernels for deblurring.
Wulff and Black~\cite{Wulff:2014:MBV} refine the latter approach and
perform segmentation into two layers, estimation of the affine
motion parameters, as well as deblurring of each layer jointly.
Relaxing the assumption of two layers and affine motion, Yamaguchi
\etal~\cite{yamaguchi2010video} and Kim and Lee~\cite{Kim:2015:GVD}
employ optical flow to approximate spatially variant blur
kernels for deblurring.
Yamaguchi \etal~\cite{yamaguchi2010video} propose deblurring based on
the flow estimates from the blurry images.
Kim and Lee~\cite{Kim:2015:GVD} iteratively refine flow estimation and
deblurred video frames by minimizing a joint energy.
The latter method represents the state-of-the-art in video
deblurring and is used for comparison in the experimental section.
To the best of our knowledge, exploiting stereo
video for deblurring has not been considered in the literature before.

Correspondence estimation on stereo video sequences can be
improved by estimating stereo correspondences and optical flow jointly
as 3D scene flow \cite{vedula1999three,quiroga2013local,wedel2011stereo}.
In our approach we build on the piecewise rigid scene flow by Vogel
\etal~\cite{Vogel:2015:3SF} for the following reasons.
First, it provides us with explicit 3D rotations and translations that
we employ for accurate blur kernel construction.
Second, through over-segmentation into planar patches, it also
delivers occlusion information, which we use as initialization for our
boundary-aware object motion deblurring.
A general problem in object motion deblurring is that object boundaries with mixed
foreground and background pixels can lead to severe ringing artifacts
(see Fig.~\ref{fig:sceneFlow}).
Explicit segmentation and $\alpha$-matting
\cite{Wulff:2014:MBV,Tai:2010:Correction} can prevent this effect, but
requires restrictive assumptions on the number of moving objects.
To handle general scenes with an arbitrary number of objects, we
extend the robust outlier handling of Chen \etal~\cite{chen2008robust}
to spatially-variant deblurring based on scene flow, and apply it to
the mixed pixels at object boundaries.

In contrast to the aforementioned deblurring approaches, 
Cho \etal~\cite{cho2012video} deblur hand-held video under the
assumption that patches are sharp in some frames of the video.
However, in the case of autonomous robots or objects passing the field
of view with high speed, this assumption does not hold.
Joshi \etal~\cite{joshi2010image} attach additional inertial measurement units to the
camera, but this does not account for
object motion.
An additional low-resolution, high frame-rate camera can provide
complex motion kernels~\cite{Tai:2010:Correction}, but does not
provide depth estimates in the way a stereo camera can.


%% file: fig_kernel.tex
\begin{figure}[t]
\centering

\subfloat[Homography-induced blur]{\label{fig:kernel:rotHom}\includegraphics[width = 0.32\columnwidth]{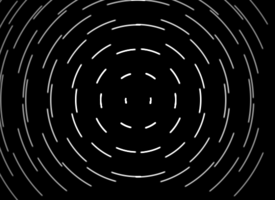} }%
\subfloat[Optical flow-induced blur]{\label{fig:kernel:rotFlow}\includegraphics[width = 0.32\columnwidth]{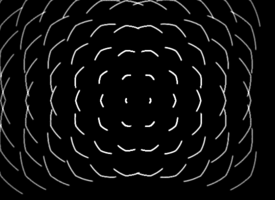} }%
\subfloat[Difference of blurs]{\label{fig:kernel:rotDiff}\includegraphics[width = 0.32\columnwidth]{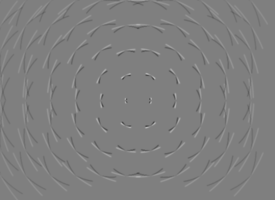} }%
\caption{Descriptiveness of homography-based blur kernels:
  Using 3D rigid body motion to generate blur kernels, we can
  faithfully express, \eg, yaw motion~\protect\subref{fig:kernel:rotHom}, while kernels constructed with spatially varying 2D displacement vectors fields~\cite{Kim:2015:GVD} only yield an
  approximation~\protect\subref{fig:kernel:rotFlow}.
  Approximation errors~\protect\subref{fig:kernel:rotDiff} are
  also present close to the rotation axis where motions are small (extremly large yaw angle and all intensities scaled for better visibility) }
\label{fig:kernel}
\end{figure}

%% file: technical.tex
\section{Blurred Image Formation in Stereo Video}
\paragraph{Inducing blur matrices from 3D rigid object motions.}
Due to the finite exposure time $\tau$ of our stereo video camera, each
frame of each camera is blurred.
Our goal is to find a sharp image $I_{t_0}$ for a reference camera at time $t_0$.
We base our approach on the scene flow of Vogel
\etal~\cite{Vogel:2015:3SF}, and likewise assume that the scene can 
be approximated with planar patches that undergo a 3D rigid body
motion.
If an object in the scene is non-planar, this assumption leads to an
over-segmentation of the object into spatially adjacent patches
(see Fig.~\ref{fig:sceneFlow:overseg}).
Considering video frames where the exposure time is naturally limited
by the frame rate, we additionally assume that the motion of each
patch is constant during the exposure time of two consecutive frames. 
Note that a constant rigid motion in 3D does
not necessarily imply that its 2D projection is constant; the projection
may, \eg in the case of a rotation, be constantly accelerated.
However, our assumption excludes rapidly changing motions such as vibrations.

Constant 3D rigid body motion can be expressed as a homogeneous $4\times4$ matrix 
\begin{equation}
  M = \begin{pmatrix} R & T \\ \vec{0} & 1 \end{pmatrix}
\end{equation}
with a rotation matrix $R\in\mathbb{R}^{3\times 3}$ and a translation
vector $T \in \mathbb{R}^3$.
To enable our highly accurate blur kernel description, we rewrite $M =
\exp\big( \theta \xi \big)$ as matrix exponential, where $\theta \in
\mathbb{R}$ describes the rotation angle and $\xi$ is a $4\times 4$
matrix that is determined by the rotation axis and the translation,
see \cite{mei2008modeling,murray1994mathematical}.
With $M$ describing the motion between time instants $t_0$ and $t_1$, the constant 
3D motion between two arbitrary time instants
$t_a$ and $t_b$ is given as
\begin{equation}
  M_{t_b, t_a} = \exp\left( \frac{t_b - t_a}{t_1 - t_0} \theta \xi \right).
\end{equation}

In a piecewise planar scene approximation, the 3D planes of the
patches at time $t$ are defined via their scaled normals $n_t$.
All points $P$ on the plane satisfy the equation $P^{\text{T}} n_t =
1$, where $P^{\text{T}}$ is the transposed of $P$.
We can relate a moving 3D point to its corresponding pixel location on
the image plane via the camera geometry.
Given the calibration matrix $K$ of the reference camera and its
location $T_K$, the projection from a 3D plane to the image plane
at time $t$ can be written in homogeneous coordinates as $Pr_{t}
= K - K T_{K} n_t^{\text{T}}$, see \eg~\cite{Vogel:2015:3SF}.

Under the assumption of color constancy, two sharp images of the reference camera (with
hypothetical infinitesimal exposure) at different times are connected via 
\begin{equation}
  I_{t_a}( x ) = I_{t_b}( \prescript{t_b}{} H^{t_a} x ) 
  \quad\text{where}\quad \prescript{t_b}{}H^{t_a}_{} =
  Pr_{t_b} M_{t_b, t_a} Pr_{t_a}^{-1}.
\end{equation}
With this notation, a blurry image pixel $x$ in the interior of a patch is formed from the reference image as
\begin{equation}\label{eq:imageFormation}
 \hat{B}(x)
 = \int^{t_0 + \frac{\tau}{2}}_{t_0 -\frac{\tau}{2}}  I_{t}(x ) \,\text{d}t
  = \int^{t_0 + \frac{\tau}{2}}_{t_0 -\frac{\tau}{2}}  I_{t_0}(  \prescript{t_0}{}H^{t}_{} x ) \,\text{d}t,
\end{equation}
where
\begin{equation}
  \prescript{t_0}{}H^{t}_{} = Pr_{t_0} \exp\big( -t \theta
  \xi \big) Pr_{t}^{-1}
\end{equation}
is a homography that can be computed exactly from camera geometry,
normal, and motion.
To put it differently, a 3D point that is projected to $x$ on the image plane 
describes a certain trajectory on the image plane during the exposure time.
If the 3D point follows a rigid body motion, 
the homography $\big( \prescript{t_0}{}H^{t}_{} \big)^{-1}$ 
allows us to exactly describe this 2D trajectory. 
In contrast, optical flow based methods~\cite{Kim:2015:GVD,Kim:2013:DSD,portz2012optical}, employ 2D optical flow vectors to generate $I_{ t}$ via forward warping. 
Thus the trajectory of a point on the image plane is approximated by a 2D line that is traversed with constant velocity. 
As optical flow is spatially variant, the trajectories may change for each pixel, hence induce blur kernels with a curved shape.
However, more complex motions such as rotations can only be
\emph{approximated}, Fig.~\ref{fig:kernel}. 
In our approach, the description of trajectories due to 3D rigid body motions is \emph{exact}.
As our experiments show, this also results in more faithfully deblurred images.

By discretizing the integration over time with $\delta t = \frac{\tau}{N}$ (we fix $N = 70$) and using bilinear
image interpolation, we can obtain a discretized version of
Eq.~\eqref{eq:imageFormation} for vectorized reference images as
$\hat{B}(x) = A_x \vec{I}_{t_0}$.
Here, $A_x$ denotes a sparse row vector that depends on the homography
estimated at pixel $x$. Stacking the blur vectors $A_x$ for each pixel, we obtain our homography-based blur matrix $A$ leading to 
$\hat{\vec{B}} = A \vec{I}_{t_0}$.

\input{tab_kernelGeneration}

\paragraph{Motion boundaries.}
If only scene points from the same plane contribute to the color $B(x)$ of the measured blurred image at point $x$, the image formation
model of Eq.~\eqref{eq:imageFormation} is exact.
If at time $t$ a scene point with a different motion contributes to $B (x)$ we should also use the corresponding homography. However, within an object, the planar patches are adjacent in space and move
consistently.
Therefore, we approximate the blur with the row vector $A_x$ induced
by the homography of $x$ at $t_0$.
At motion boundaries, the homographies are very different 
and as pixels of foreground and background mix, transparency effects occur.
While such effects can be modeled, taking them into
account requires precise localization of the motion boundaries,
which is very challenging.
Instead, we exclude motion boundaries from the deblurring process by
means of an iterative approach. 
In each iteration, we downweight pixels with a high difference between image formation
model and measured image and try to find a sharp image that explains the remaining pixels.
Under the assumption of additive Gaussian noise, we use the residual
to compute a weight for each pixel as 
\begin{equation}\label{eq:occWeight}
w_n  (x) = \exp\Big( - k_\sigma \| B (x) - A_x \vec{I}^{n-1}_{t_0}  \|^2 \Big),
\end{equation}
where $\vec{I}^{n-1}_{t_0} $ denotes the current estimate of the
sharp (color) image.
For normalized images we set $k_\sigma = \nicefrac{4000}{3}$ as default value.
In the first iteration we initialize $w_0 $ with the binary
occlusion information from the scene flow.
As Fig.~\ref{fig:tech} shows, the weights converge quickly.
Some pixels in the image that were initially suppressed as motion
boundaries are included in deblurring at a later iteration.
More importantly, other pixels where the image formation model is
invalid are suppressed later on, which helps controlling ringing
artifacts.
Suppression may also happen due to some inaccuracies in the computed scene flow.
In the experimental section, we will see how this property actually
helps to improve deblurring results. 
%

\input{fig_technical}

\paragraph{Deblurring.}
Theoretically, we could fill in the regions at motion boundaries during
deblurring by using adjacent frames or information from the other camera.
However, we found experimentally that correspondence estimation in
these regions is too unreliable to produce visually pleasing results.
Instead, we exploit that natural, sharp images follow a Laplacian
distribution of their gradients~\cite{Levin:2006:BMD}.
In locations where the image formation model is unreliable, \eg, at
motion boundaries, we rely on this prior to provide the necessary
regularization.
Specifically, we obtain an estimate of the sharp reference frame by
minimizing the energy 
\begin{equation}\label{eq:IRLS}
 E( \vec{I}_{t_0}  ) = \sum_{x\in\Omega}
 \Big\| w_n (x) \big( B (x) - A_x \vec{I}_{t_0}  \big) \Big\|^2 + \alpha \rho\big( \nabla I_{t_0} (x) \big),
\end{equation}
where $\Omega \subset \mathbb{N}^2$ is the image domain and the constant $\alpha$ is fixed to $0.001$.
Following prior work~\cite{Levin:2006:BMD}, we use the robust norm
$\rho\big( c \big) = | c | ^{0.8}$ for each color channel and gradient direction.
To solve the optimization problem
 in \Eq~\eqref{eq:IRLS}, we use iteratively reweighted least squares
(IRLS)~\cite{Levin:2007:UAS}. 
In each reweighting iteration, we compute the following weights 
\begin{equation}
\rho_n(c) = \frac{1}{c} \frac{d \rho\big(c\big) }{d c } 
\approx \max\big( |c |, \epsilon  \big)^{0.8-2} \quad \text{ with } \quad \epsilon = 0.01
\end{equation}
for
the smoothness term using the preceding
image estimate $\nabla I^{n-1}_{t_0} $.
Then we minimize the least squares energy 
\begin{equation}
E( \vec{I}_{t_0} , n ) = \sum_{x\in\Omega} 
\big\| w_n (x) \big( B (x) - A_x \vec{I}^n_{t_0}  \big) \big\|^2 + \alpha \| \rho_n \nabla I^n_{t_0}  (x) \|^2
\end{equation}
via conjugate gradients.
We alternate between updating the occlusion weight $w_n $
and the smoothness weight $\rho_n$.
In all our experiments the weights converge quickly and only a few
($\approx 10$) iterations were needed in total.
To compute the 3D scene flow needed for our stereo video deblurring
approach, we rely on the method of Vogel \etal~\cite{Vogel:2015:3SF}.
The algorithm is originally designed for sharp images.
However, its data term uses the
census transform for comparing the warped images, which makes it quite
robust to image blur.
Of course, scene flow estimation will reach its limits for very strong motion blur.
Experimentally, we find that by aggregating evidence in piecewise
planar patches, the method yields a scene flow accuracy 
that turns out to work well in deblurring stereo videos of casual motion.
As the following experiments will show, it is crucial, however, to not
only rely on the robust correspondence information, but to exploit the
homographies to directly induce the blur kernels.


%% file: tab_kernelGeneration.tex
\begin{table}[bt]
  \caption{
    Overview of the different sources of motion information used for
    video deblurring:
    When pure 2D correspondence is considered (top two rows), the
    induced blur kernels are only approximate, as motion trajectories
    are assumed to be linear.
    Exploiting homographies from scene flow allows us to capture the
    fact that rigid 3D object motion leads to non-linear trajectories
  }
\label{tab:kernel}
\centering
\smallskip
\begin{tabu} to \linewidth{@{}lX[1,c]X[1,c]X[1.2,c]@{}}
\toprule
& & & linear approximation \\
Motion information  & \# of frames & \# of views & of trajectories\\
\midrule
optical flow & 2 & 1 & yes \\
2D projection of scene flow & 2 & 2 & yes\\
homographies & 2 & 2 & no \\
\bottomrule
\end{tabu}

\end{table}

%% file: fig_technical.tex
\begin{figure}[t]
\centering
\subfloat[Input]{\label{fig:occlusion:chair}\includegraphics[width = 0.25\columnwidth]{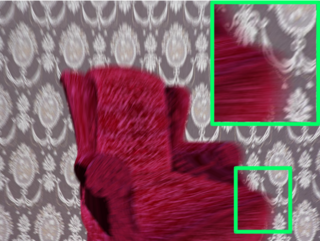} }%
\subfloat[Initial]{\label{fig:occlusion:init}\includegraphics[width = 0.25\columnwidth]{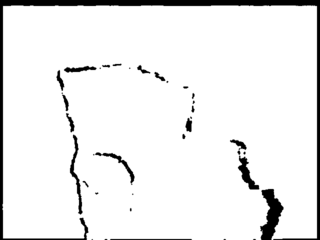} }%
\subfloat[Iteration 3]{\label{fig:occlusion:mid}\includegraphics[width = 0.25\columnwidth]{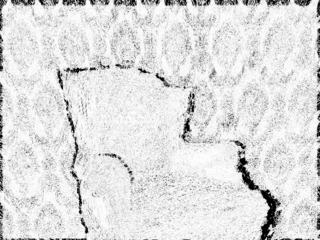} }%
\subfloat[Iteration 10]{\label{fig:occlusion:final}\includegraphics[width = 0.25\columnwidth]{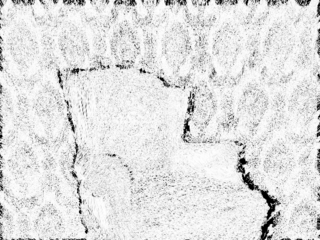} }%
\caption{Downweighting of mixed pixels due to motion boundaries:
  Foreground and background mix at motion
  boundaries and violate our image formation model~\protect\subref{fig:occlusion:chair}.
  At motion boundaries and locations of inaccurate flow estimates,
  the image formation model is downweighed to avoid
  ringing artifacts.
  We initialize these weights with the occlusion information
  provided by the scene flow \protect\subref{fig:occlusion:init} and
  refine them
  iteratively~\protect\subref{fig:occlusion:mid},\protect\subref{fig:occlusion:final}}
\label{fig:tech}
\end{figure}

%% file: experiments.tex
\section{Experiments}
\input{fig_planar}
To demonstrate the efficacy of the proposed stereo video deblurring,
we perform experiments on synthetic images with known ground truth,
as well as on real images. 
We capture the real video footage with a \emph{Point Grey Bumblebee2}
stereo color camera, which can acquire $640 \times 480$ images at a
frame rate of $20$ Hz.
We use the internal calibration and supplied software to obtain
rectified and demosaiced images. 
The exposure time of each image can be obtained from the camera
software.
In all experiments, we compute scene flow using the publicly available
implementation of \cite{Vogel:2015:3SF}.
We take the default parameters and scale them uniformly to account for
the baseline difference between our stereo camera and the KITTI
dataset \cite{Geiger:2012:AWR} for which they were tuned. 
For the $640\times 480$ image in Fig.~\ref{fig:sceneFlow} our approach
requires \SI{73}{\second} to form the discretized blur matrix $A$.
Using MATLAB to optimize Eq.~\eqref{eq:IRLS} in $25$ conjugate gradient
steps and $10$ IRLS iterations requires \SI{69}{\second} on an
$8$-core \SI{4}{\giga\hertz} CPU.

\input{tab_planar}

\subsection{Comparing flow-based deblurring to homography-based deblurring}
We begin by applying the proposed stereo video deblurring to scenes
without object discontinuities.
In this way we can analyze the benefit of the
homography-induced motion blur model in isolation.
We create synthetic sequences by simulating various 3D motions (upward
and forward translation, and a combination forward translation and
yaw) of a planar, roughly fronto-parallel texture, see~Fig.~\ref{fig:planar:input}.
\footnote{More textures and motions are
  evaluated in the supplemental material.}
A second test set consists of rigidly moving 3D objects rendered with a raytracer at very small time steps and averaged to give motion-blurred images (see Figs.~\ref{fig:occlusion:chair},~\ref{fig:saveMask:input} and~\ref{fig:ray:input} for the first image of the left view).
We take the central frame of each motion-blurred image as a sharp
reference frame.
For the rendered scenes motion discontinuities are known. 
In the first experiment, we disable the data term around any motion
discontinuities by fixing the weights $w_n$ in these areas to
zero, see Fig.~\ref{fig:saveMask:mask} for an example.
As the image prior stays active, the boundaries are filled in smoothly
as illustrated in Fig.~\ref{fig:saveMask:3DSF}.
We compare our homography-induced deblurring approach against
deblurring with blur matrices generated from different 2D displacement
fields.
We use forward and backward 2D motion as described by Kim and Lee~\cite{Kim:2015:GVD} and apply them in our IRLS deblurring framework.
In particular, we use the known \emph{ground truth 2D displacement},
the \emph{2D initial optical flow} with which the scene flow is
initialized~\cite{Vogel:2013:EDC} ( \emph{baseline deblurring} ), and the \emph{2D projection of the
  scene flow} to induce blur kernels.
Table~\ref{tab:kernel} summarizes these settings.
Table~\ref{tab:planar} shows the peak-signal-to-noise-ratio (PSNR) of
the deblurred images from the different methods.
We observe that the PSNR of our homography-based stereo video deblurring outperforms the results of deblurring with ground truth 2D displacement in all cases of non-fronto-parallel motion.
In these cases linear motion trajectories of constant velocity are an
approximation.
Blur matrices induced by
homographies are more expressive and improve the results.
Already, deblurring with the 2D projection of scene flow achieves a
consistently higher PSNR than deblurring with the initial flow.
Indeed, in the case of forward motion, also deblurring with the 2D projection of the scene flow outperforms deblurring with ground truth displacement.
The estimated 2D displacement appears to be a better approximation to
the linear, but accelerated trajectory of the 3D forward motion than
the 2D ground truth displacement. 
Figs.~\ref{fig:planar:GT}~and~\ref{fig:planar:3SF} show examples of
deblurred images using the ground-truth 2D displacement and our
homography-based approach.
From the difference image between the results and the original sharp
texture, Figs.~\ref{fig:planar:GTdiff}~and~\ref{fig:planar:3SFdiff}, we observe that the increase in PSNR is due to the mitigation of ringing effects throughout the image.

For the raytraced scenes the geometry of the moving objects is non-planar and the planarity assumption in our image formation model becomes an approximation.
Fig.~\ref{fig:saveMask} shows the estimated disparity of an object and the deblurred image obtained by masking out discontinuities.
Looking at the difference image, Fig.~\ref{fig:saveMask:3SFdiff}, we observe that the deblurring error
for slightly curved surfaces is comparable to the performance on
planar regions of the background, showing that the
over-segmentation aids coping with curved surfaces.
For all rendered scenes where the disparity does not exhibit gross
errors, we observe in Table~\ref{tab:planar} that 3D homography-based
deblurring improves the PSNR clearly over any form of 2D deblurring.
In the scene `\emph{apples}', Fig.~\ref{fig:ray:input}, 1st row, depth estimation fails with a mean disparity error of $4.95$ pixels.
In this situation the deblurring quality of homography-based
deblurring drops below that of its 2D projection. 
Still, both outperform the results obtained with the initial optical flow.
More importantly, as we will see below, the iterative weighting scheme
for treating motion discontinuities can address such disparity
estimation errors as well and lead to much improved results.

\input{fig_raytrace}

\subsection{Full algorithm with motion discontinuities}
\input{tab_occlusion}
We now evaluate the performance of stereo video deblurring in the
presence of object motion boundaries. 
We use the raytraced scenes from the previous experiment, but this
time without providing ground-truth information on the motion
discontinuities, Fig.~\ref{fig:ray:input}.
Additionally, we use real images captured with a stereo camera
attached to a motorized rail, Fig.~\ref{fig:avrg:input}.
The camera moves forward very slowly on the rail while we capture
frames with maximal exposure time and frame rate.
By averaging the frames, we obtain motion-blurred images.
Comparison to the central frame of the averaged frame series allows for
numerical evaluation.
Finally, we capture scenes with arbitrarily moving objects for which
only a visual evaluation is possible, Fig.~\ref{fig:real:input}.
As before we compare against 2D versions of our algorithm.
Additionally, we compare against the state-of-the-art video deblurring algorithm of Kim and Lee~\cite{Kim:2015:GVD} that uses $3$ consecutive monocular frames.
We tuned their regularization parameter to obtain the most accurate results.
%

In Fig.~\ref{fig:ray:3Dall} and Fig.~\ref{fig:ray:3SF} we first contrast
homography-induced deblurring without and with handling of motion
boundaries.
When not taking into account motion boundaries explicitly, \ie $w_n \equiv 1$,
Fig.~\ref{fig:ray:3Dall}, considerable ringing artifacts are the
result, but they are successfully suppressed with our proposed
iterative weighting scheme, Fig.~\ref{fig:ray:3SF}.
This also becomes evident in the numerical evaluation when comparing
the \nth{3} and \nth{4} column of Tab.~\ref{table:occlusion}~(top)
\footnote{In the supplemental material we also consider the Structural Similarity index \cite{wang2004image} to compare the results.}.
For the real sequences in Fig.~\ref{fig:avrg}, boundary artifacts are
generally less pronounced, as all objects in the scene are static and
the camera moves toward the scene. 
However, as shown in Fig.~\ref{fig:tech}, the discontinuity weight can
still compensate errors in scene flow computation.
One such example is the erroneous depth estimation in the
`\emph{apples}' scene, which is disabled by the discontinuity weight.
Similarly, also in the scenes with the motorized rail, our full
object motion deblurring approach improves the PSNR compared to the
basic homography approach, Tab.~\ref{table:occlusion}~(bottom).

When comparing to the
state-of-the-art video deblurring method of Kim and Lee
\cite{Kim:2015:GVD}, we find that our stereo video deblurring approach
yields significantly fewer ringing artifacts and considerably sharper
results.
This can be seen visually, comparing (d) to (c) of
Figs.~\ref{fig:ray}~to~\ref{fig:reals}, as well as quantitatively in
Table~\ref{table:occlusion}.
Interestingly, we find in Table~\ref{table:occlusion} that IRLS
deblurring with the 2D projection of the scene flow is already on par
with video deblurring of Kim and Lee.
3D homography-based deblurring without boundary handling improves on these
result numerically already, highlighting the importance of our
homography-induced blur kernels.
Yet, our full homography-based object deblurring with motion boundary
handling gives further numerical gains and a large visual
improvement.
Recall that the motion boundaries are initially obtained from the 3D
scene flow, thus unique to our setting.

For the real scenes with independent object motion,
Fig.~\ref{fig:reals}, we observe that the optical flow-based
approaches introduce ringing artifacts, particularly where strong
gradients of the background coincide with the object boundary.
Our stereo video deblurring algorithm can cope with this situation
even in the presence of non-planar, non-rigidly moving objects such as the trousers
(\nth{2} row) are present.

\input{fig_average}
\input{fig_reals}


%% file: fig_planar.tex
\begin{figure}[tb]
\centering
%
%
\subfloat[Input]{\label{fig:planar:input}
\includegraphics[width = 0.19\columnwidth]{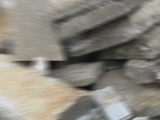} }%
\subfloat[PSNR $30.03$]{\label{fig:planar:GT}\includegraphics[width = 0.19\columnwidth]{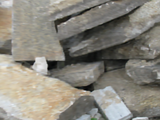} }%
\subfloat[Difference~(b) to\newline reference image]{\label{fig:planar:GTdiff}\includegraphics[width = 0.19\columnwidth]{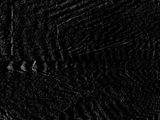} }%
\subfloat[PSNR $31.02$]{\label{fig:planar:3SF}\includegraphics[width = 0.19\columnwidth]{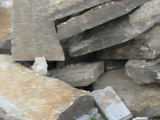} }%
\subfloat[Difference~(d) to\newline reference image]{\label{fig:planar:3SFdiff}\includegraphics[width = 0.19\columnwidth]{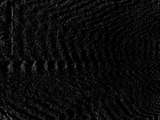} }%
%
%
\caption{Deblurring planar textures: For a planar texture blurred with 3D rigid body
  motion~\protect\subref{fig:planar:input}, deblurring with 2D
  spatially-variant ground truth displacements~\protect\subref{fig:planar:GT}
  yields ringing errors~\protect\subref{fig:planar:GTdiff} that can be reduced by
  deblurring with our homography-based image formation model~\protect\subref{fig:planar:3SF},~\protect\subref{fig:planar:3SFdiff}
}
\label{fig:planar}
\end{figure}

\begin{figure}
\centering
%
\subfloat[Input]{\label{fig:saveMask:input}
\includegraphics[width = 0.30\columnwidth]{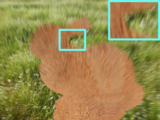} }%
\subfloat[Mask]{\label{fig:saveMask:mask}\includegraphics[width = 0.30\columnwidth]{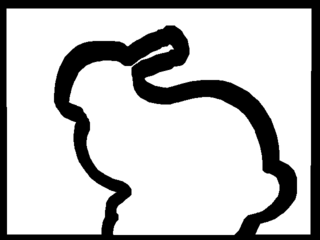} }%
\subfloat[Disparity]{\label{fig:saveMask:disp}\includegraphics[width = 0.38\columnwidth]{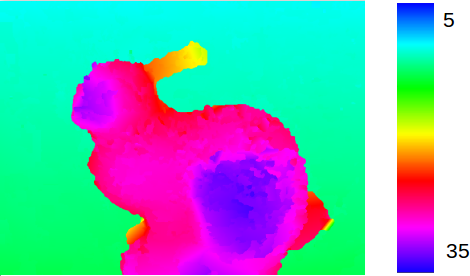} }%
\\
\hspace{-0.8cm}
\subfloat[Masked deblurring]{\label{fig:saveMask:3DSF}\includegraphics[width = 0.30\columnwidth]{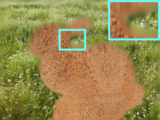} }%
\subfloat[Difference (d) to (f) on (b)]{\label{fig:saveMask:3SFdiff}\includegraphics[width = 0.30\columnwidth]{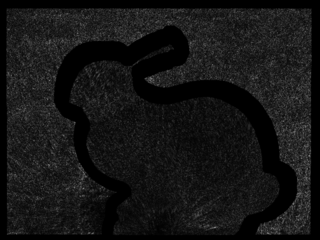} }%
\subfloat[Sharp reference image]{\label{fig:saveMask:GTimg}\includegraphics[width = 0.30\columnwidth]{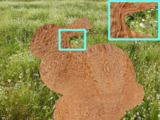} }%
%
%
\caption{Deblurring with masked discontinuities: Our raytraced stereo
  video frames contain independent object motion of non-planar
  objects~\protect\subref{fig:saveMask:input}. Through the estimated
  disparity we can assess the shape of the
  objects~\protect\subref{fig:saveMask:disp}. Excluding given
  discontinuities~\protect\subref{fig:saveMask:mask} from the
  computation of the data term, invalid areas are filled in
  smoothly~\protect\subref{fig:saveMask:3DSF}. The masked difference
  image~\protect\subref{fig:saveMask:3SFdiff} to the real sharp
  image~\protect\subref{fig:saveMask:GTimg} shows that
  homography-based deblurring has about the same error in planar as in
  curved surfaces, showing the effectiveness of the over-segmentation
  from the scene flow
}
\label{fig:saveMask}
\end{figure}

%% file: tab_planar.tex
\begin{table}[bt]
\caption{Deblurring without considering motion-discontinuity regions: 
For different motions of a planar texture (top) and
  moving 3D objects with masked object boundaries (bottom),
  we report the peak signal-to-noise ratio (PSNR) of the deblurred
  reference frame, the average endpoint error of the estimated motion
  (AEP), and the average disparity error (ADE) of the estimation.  
  For all scenes the use of scene flow increases deblurring accuracy
  compared to using optical flow.
  For scenes with non-fronto-parallel motion (all except `\emph{upward}' and
  `\emph{apples}') homography-based object motion deblurring provides the best
  results (bold)} 
\label{tab:planar}
\smallskip
\centering
\setlength{\tabcolsep}{8pt}
\begin{tabu}{@{}l*7{c}@{}}
\toprule
blur kernel & \multicolumn{1}{c}{ground truth} &
\multicolumn{2}{c}{initial optical} & \multicolumn{2}{c}{2D projection}  & \multicolumn{2}{c@{}}{3D}\\
source & \multicolumn{1}{c}{2D displacement} & \multicolumn{2}{c}{flow \cite{Vogel:2013:EDC}} & \multicolumn{2}{c}{of scene flow}  & \multicolumn{2}{c@{}}{homographies}
\\
\cmidrule(lr){2-2}
\cmidrule(lr){3-4}
\cmidrule(lr){5-6}
\cmidrule(lr){7-8}
& PSNR & AEP & PSNR & AEP & PSNR & ADE & PSNR\\
 \midrule
upward & {\bf 26.09} & 2.53 & 24.89& 0.20 & 25.83 & 3.66 & 25.94\\ 
forward & 34.12 & 0.09 & 34.45& 0.10 & 34.55 & 0.19 & {\bf 34.74}\\
forward + yaw & 30.03 & 0.15& 29.90 & 0.15 & 29.95 & 0.45 & {\bf 31.02}\\
\hline
apples            &  34.36& 0.48& 29.39 & 0.81& {\bf34.74} &4.95& 33.33 \\
bunny            &  25.01& 0.62& 24.87 & 0.54& {25.09} &0.50& {\bf25.19} \\
chair            &  24.84& 0.59& 23.86 & 0.47& {24.81}      &0.25& {\bf25.03} \\
squares          &  25.97& 2.32& 22.69 & 0.97& {25.82}      &0.65& {\bf27.21} \\
triplane         &  27.43& 1.16& 27.11 & 0.49& {27.27}      &0.12& {\bf28.30} \\
\bottomrule
\end{tabu}

\end{table}
\setlength{\tabcolsep}{1.4pt}

%% file: fig_raytrace.tex
\begin{figure}[t]
\centering
\hspace{-0.25cm}
\rotatebox{90}{\hspace{0.72cm} apples}{\includegraphics[width = 0.24\columnwidth]{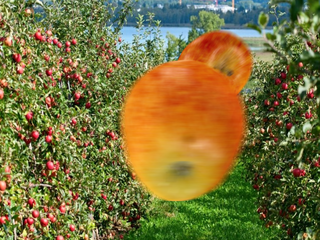} }%
%
{\includegraphics[width = 0.24\columnwidth]{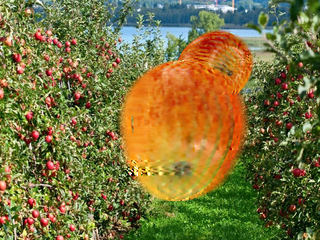} }%
%
{\includegraphics[width = 0.24\columnwidth]{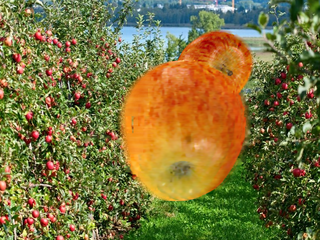} }%
%
{\includegraphics[width = 0.24\columnwidth]{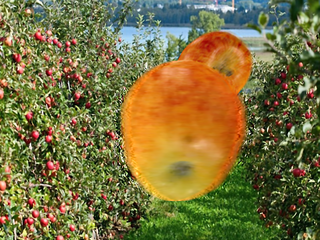} }%
\\
\hspace{-0.25cm}
\rotatebox{90}{\hspace{0.72cm} bunny}{\includegraphics[width = 0.24\columnwidth]{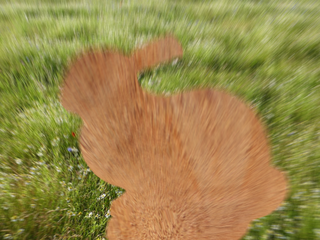} }%
%
{\includegraphics[width = 0.24\columnwidth]{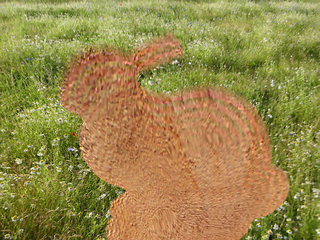} }%
%
{\includegraphics[width = 0.24\columnwidth]{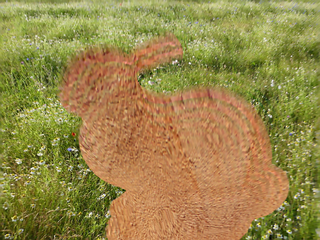} }%
%
{\includegraphics[width = 0.24\columnwidth]{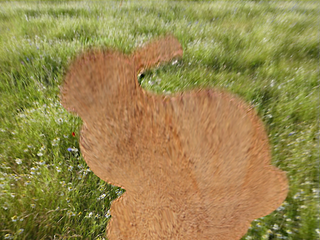} }%
\\
\hspace{-0.2cm}
\rotatebox{90}{\hspace{0.72cm} squares}{\includegraphics[width = 0.24\columnwidth]{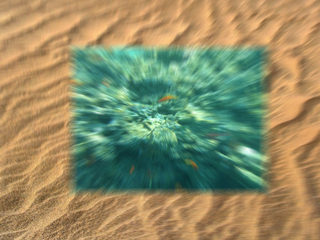} }%
%
{\includegraphics[width = 0.24\columnwidth]{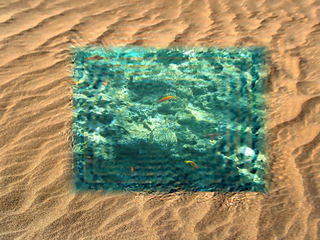} }%
%
{\includegraphics[width = 0.24\columnwidth]{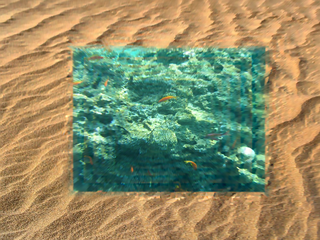} }%
%
{\includegraphics[width = 0.24\columnwidth]{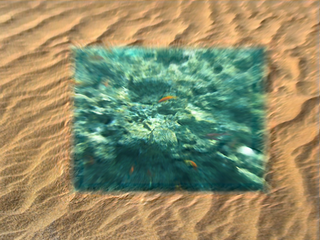}}%
\\
\vspace{-0.35cm}
\hspace{-0.12cm}%
\rotatebox{90}{\hspace{0.72cm} triplane}%
\subfloat[Input]{\label{fig:ray:input}\includegraphics[width = 0.24\columnwidth]{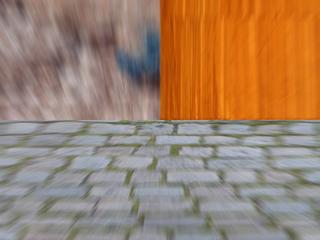} }%
\subfloat[Ours (no boundary downweighting)]{\label{fig:ray:3Dall}\includegraphics[width = 0.24\columnwidth]{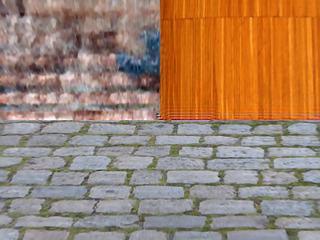} }%
\subfloat[Ours (full)]{\label{fig:ray:3SF}\includegraphics[width = 0.24\columnwidth]{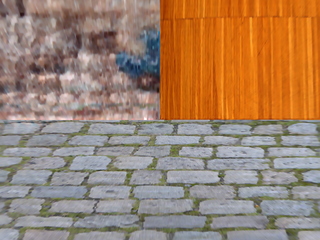} }%
\subfloat[Kim and Lee~\cite{Kim:2015:GVD}]{\label{fig:ray:kim}\includegraphics[width = 0.24\columnwidth]{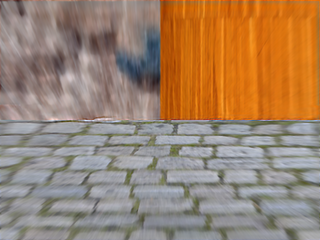} }%
\\
\caption{Raytraced scenes for evaluating object motion deblurring: The
  input images exhibit blur due to 3D object
  motion~\protect\subref{fig:ray:input}. Also when 3D homographies are
  used to induce blur kernels, mixed pixels at object boundaries cause
  some ringing artifacts~\protect\subref{fig:ray:3Dall}.
  Iteratively downweighting the boundaries from the data term, our
  full stereo video deblurring~\protect\subref{fig:ray:3SF} suppresses ringing and obtains
  considerably sharper images than state-of-the-art
  video deblurring~\protect\subref{fig:ray:kim}. Please zoom in for
  detail
}
\label{fig:ray}
\end{figure}

%% file: tab_occlusion.tex
\setlength{\tabcolsep}{6pt}
\begin{table}[bt]
\caption{Deblurring with motion discontinuities:
PSNR of deblurred synthetic scenes with motion discontinuities (top) and real scenes with the camera moving on a motorized rail (bottom).
Our homography-based stereo video deblurring with motion boundary weighting (full) clearly outperforms monocular video deblurring with optical-flow induced blur kernels in all cases}
\label{table:occlusion}
\centering
\setlength{\tabcolsep}{9pt}
\begin{tabu}{@{}lccccccccc@{}}
\toprule
& \multicolumn{1}{c}{initial optical} & \multicolumn{1}{c}{2D projection} & \multicolumn{1}{c}{3D} & \multicolumn{1}{c}{ours}  & \multicolumn{1}{c@{}}{}
\\
& \multicolumn{1}{c}{flow \cite{Vogel:2013:EDC}} & \multicolumn{1}{c}{of scene flow} & \multicolumn{1}{c}{homographies} & \multicolumn{1}{c}{(full)}  & \multicolumn{1}{c@{}}{Kim and Lee \cite{Kim:2015:GVD}}\\
\midrule
apples           &20.89&  29.43& 26.00 &{\bf 29.48}& 26.14 \\
bunny            &20.72&  22.36& 22.95 &{\bf 23.20}& 21.93 \\
chair            &19.03&  21.57& 22.19 &{\bf 23.36}& 21.78 \\
squares          &19.60&  21.90& 22.56 &{\bf 24.58}& 22.99 \\
triplane         &24.73&  24.55& 25.22 &{\bf 26.59}& 23.30 \\
\hline
bottles          &29.51&  29.55& 30.80 &{\bf 31.07}& 28.33 \\
office           &31.37&  30.87& 32.45 &{\bf 32.56}& 29.01 \\
planar           &30.71&  31.21& 32.33 &{\bf 32.82}& 30.01 \\
toys             &33.51&  33.64& 34.89 &{\bf 34.90}& 31.06 \\
\bottomrule
\end{tabu}

\end{table}
\setlength{\tabcolsep}{1.4pt}

%% file: fig_average.tex
\begin{figure}[t]
\centering
\hspace{-0.25cm}
\rotatebox{90}{\hspace{0.72cm} bottles~{\color{white}q}}%
{\includegraphics[width = 0.24\columnwidth]{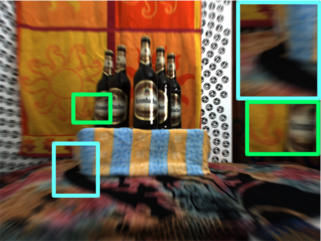} }%
%
{\includegraphics[width = 0.24\columnwidth]{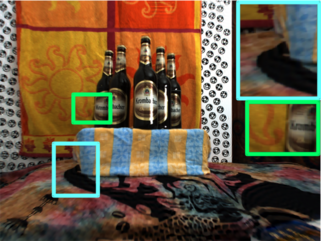} }%
%
{\includegraphics[width = 0.24\columnwidth]{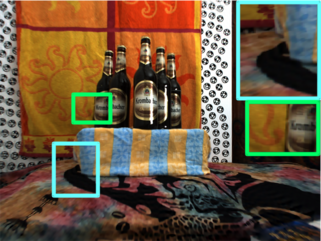} }%
%
{\includegraphics[width = 0.24\columnwidth]{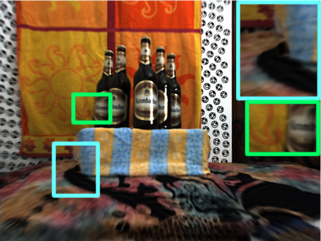} }%
\\
\hspace{-0.25cm}
\rotatebox{90}{\hspace{0.72cm} office~{\color{white}q}}%
{\includegraphics[width = 0.24\columnwidth]{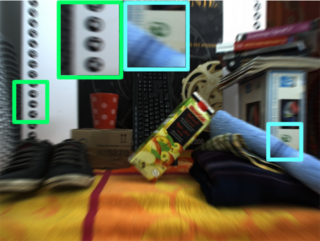} }%
%
{\includegraphics[width = 0.24\columnwidth]{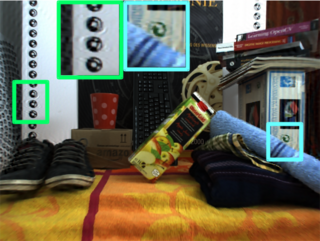} }%
%
{\includegraphics[width = 0.24\columnwidth]{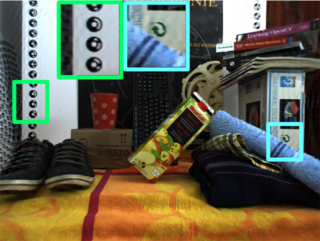} }%
%
{\includegraphics[width = 0.24\columnwidth]{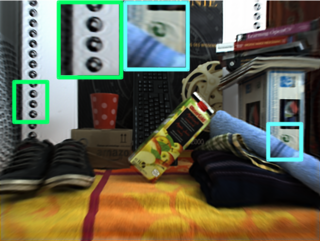} }%
\\
\hspace{-0.25cm}
\rotatebox{90}{\hspace{0.72cm} planar}%
{\includegraphics[width = 0.24\columnwidth]{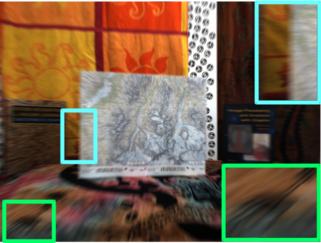} }%
%
{\includegraphics[width = 0.24\columnwidth]{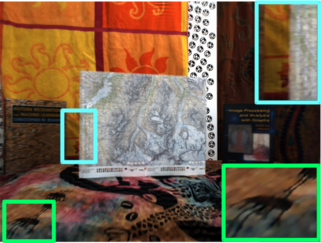} }%
%
{\includegraphics[width = 0.24\columnwidth]{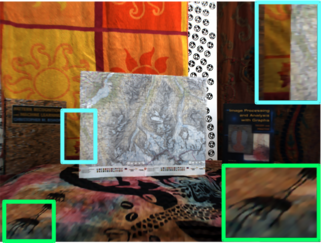} }%
%
{\includegraphics[width = 0.24\columnwidth]{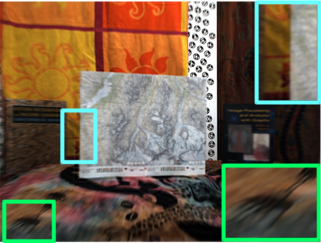} }%
\\
\vspace{-0.35cm}
\hspace{-0.12cm}%
\rotatebox{90}{\hspace{0.72cm} toys}%
\subfloat[Input]{\label{fig:avrg:input}\includegraphics[width = 0.24\columnwidth]{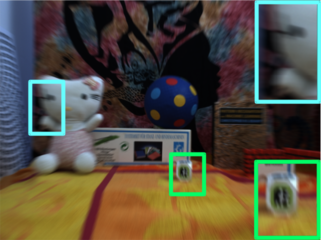} }%
\subfloat[Baseline]{\label{fig:avrg:TVAll}\includegraphics[width = 0.24\columnwidth]{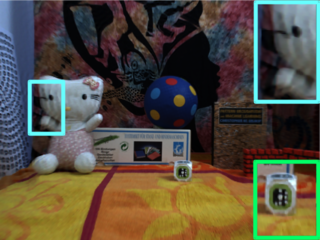} }%
\subfloat[Ours (full)]{\label{fig:avrg:3SF}\includegraphics[width = 0.24\columnwidth]{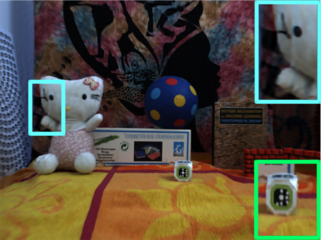} }%
\subfloat[Kim and Lee~\cite{Kim:2015:GVD}]{\label{fig:avrg:kim}\includegraphics[width = 0.24\columnwidth]{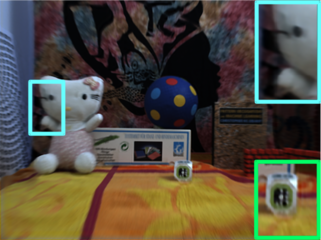} }%
\caption{Controlled camera motion for evaluating object motion deblurring: Our 3D deblurring~\protect\subref{fig:avrg:3SF} has less ringing artifacts than baseline deblurring with optical flow~\protect\subref{fig:avrg:TVAll}, and sharper results than video deblurring~\protect\subref{fig:avrg:kim}, in particular at the periphery of the images where motion is large}
\label{fig:avrg}
\end{figure}

%% file: fig_reals.tex
\begin{figure}[t]
\centering
\hspace{-0.2cm}
{\includegraphics[width = 0.24\columnwidth]{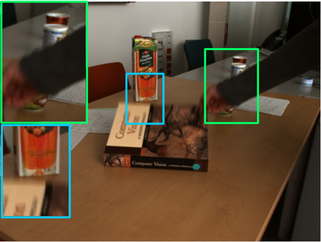} }%
%
{\includegraphics[width = 0.24\columnwidth]{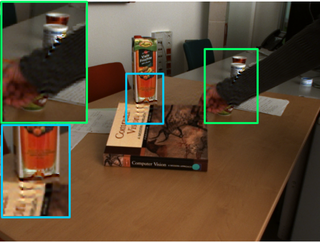} }%
%
{\includegraphics[width = 0.24\columnwidth]{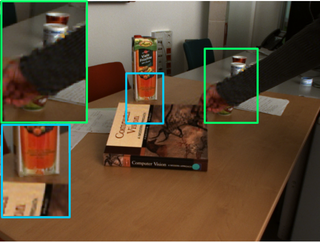} }%
%
{\includegraphics[width = 0.24\columnwidth]{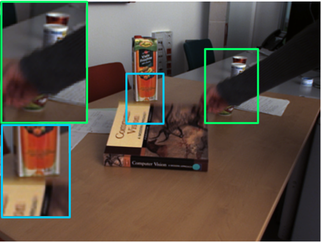} }%
\\
\hspace{-0.2cm}
{\includegraphics[width = 0.24\columnwidth]{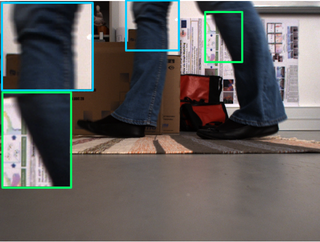} }%
%
{\includegraphics[width = 0.24\columnwidth]{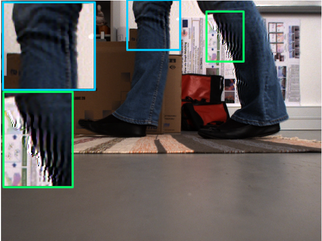} }%
%
{\includegraphics[width = 0.24\columnwidth]{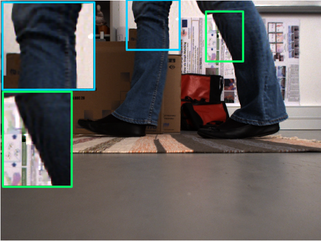} }%
%
{\includegraphics[width = 0.24\columnwidth]{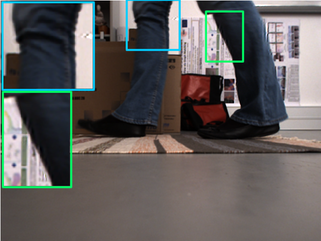} }%
\\
\hspace{-0.2cm}
{\includegraphics[width = 0.24\columnwidth]{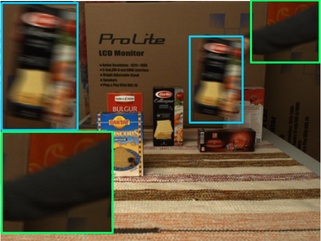} }%
%
{\includegraphics[width = 0.24\columnwidth]{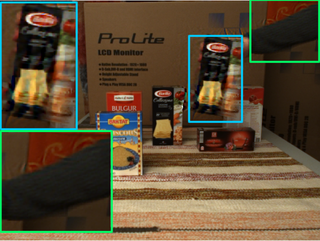} }%
%
{\includegraphics[width = 0.24\columnwidth]{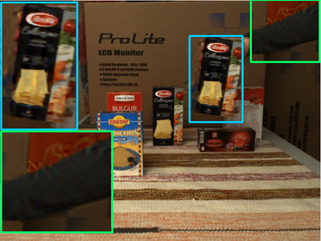} }%
%
{\includegraphics[width = 0.24\columnwidth]{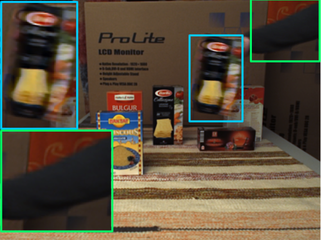} }%
\\
\vspace{-0.35cm}
\subfloat[Input]{\label{fig:real:input}\includegraphics[width = 0.24\columnwidth]{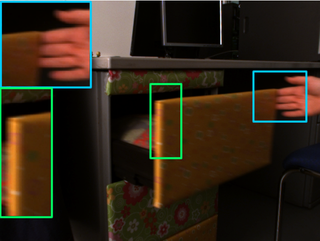} }%
\subfloat[Baseline]{\label{fig:real:TVAll}\includegraphics[width = 0.24\columnwidth]{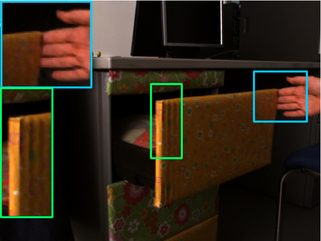} }%
\subfloat[Ours (full)]{\label{fig:real:3SF}\includegraphics[width = 0.24\columnwidth]{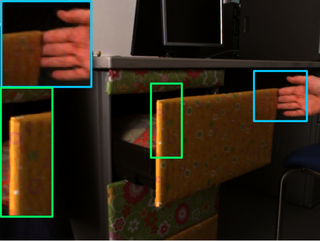} }%
\subfloat[Kim and Lee\cite{Kim:2015:GVD}]{\label{fig:real:kim}\includegraphics[width = 0.24\columnwidth]{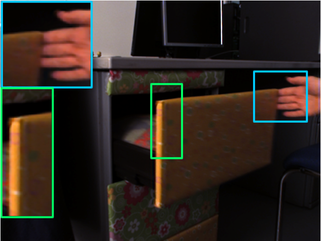} }%
\caption{For real scenes with independent object
  motion~\protect\subref{fig:real:input}, our novel stereo video
  deblurring approach~\protect\subref{fig:real:3SF} generates fewer ringing artifacts due to object boundaries than baseline deblurring with optical flow~\protect\subref{fig:real:TVAll} and sharper images than video deblurring~\protect\subref{fig:real:kim}}
\label{fig:reals}
\end{figure}

%% file: conclusion.tex
\section{Conclusions and Future Work}
We have proposed the first stereo video deblurring
approach, which is based on an image formation model that exploits 
3D scene flow computed from stereo video. 
For scenes with an arbitrary number of moving objects, we use an
over-segmentation of the scene into planar patches to establish
spatially-variant blur matrices based on local homographies.
Our experiments on synthetic scenes and real videos show that
deblurring with these homographies is more accurate than baseline
methods based on 2D linear motion approximations, as well as
the current state-of-the-art in video deblurring. 
Combined with our robust treatment of motion boundaries through an
iterative weighting scheme, our approach obtains superior results also
on real stereo videos with independently moving objects.
In future work we would like to improve the performance of scene flow
computation at motion boundaries such that we can benefit from
another view to supply information near motion boundaries.
%
{\small
\subsubsection*{\bf Acknowledgements.}
The research leading to these results has received funding from the European Research Council under the European Union's Seventh Framework Programme (FP7/2007 - 2013) / ERC Grant Agreement No. 307942 and under the European Union's Horizon 2020 research and innovation programme / ERC Grant Agreement No. 647769.
}

%% file: arxiv_supplement.tex
%
\title{Stereo Video Deblurring} 
\subtitle{ -- Supplemental Material -- }
\titlerunning{Stereo Video Deblurring}
%
%

\author{Anita Sellent\inst{1,2}  \and Carsten Rother\inst{1} \and Stefan Roth\inst{2}}
\institute{Technische Universit\"at Dresden, Germany \and Technische Universit\"at Darmstadt, Germany }
\maketitle

\renewcommand{\thesection}{\Alph{section}}
\setcounter{subsection}{0}%
\setcounter{figure}{9}
\setcounter{table}{3}
\setcounter{equation}{9}
\setcounter{footnote}{0}%

\section{Details on Homography-Based Blur Kernels}
One of the key contributions in our stereo video deblurring is to
employ 3D scene flow to induce blur kernels based on homographies.
As the difference to inducing blur kernels from an optical flow field
may seem subtle but increases deblurring performance considerably, we
schematically illustrate the difference of the two ways of generating
blur kernels in Fig.~\ref{fig:pptKernels}.
\input{fig_pptKernels}

\section{Over-segmentation of Non-Planar Scenes}
\input{fig_sceneFlow2}
To enable the use of homographies, we assume that the scene consist of planes.
For objects of arbitrary shape, we approximate their geometry with a collection of planar patches.
Figure \ref{fig:sceneFlow2:overseg} shows the piecewise planar approximations for real scenes.
In all our experiments we obtained approximations with $1000-1500$
planar patches of different 3D motions for images of size $640\times
480$ pixels. 
\section{Deblurring Planar Patches}
In the main paper we show that homography-based deblurring improves
the PSNR for a planar texture undergoing 3 different 3D motions.
Here we provide additional results using other textures and types of
motions to show that the improvement generalizes.
Fig.~\ref{fig:planar2} shows the different planar textures undergoing various 3D motions.
The difference images
(Figs.~\ref{fig:planar2:GTdiff} and \ref{fig:planar2:3SFdiff}) show that in all cases
homography-based deblurring removes subtle errors that optical
flow-based kernel approximation accumulates all over the texture.
Additionally, some localized error of the initial optical flow estimation is also corrected in our stereo video deblurring (\eg rows 2, 4, 6).
We average the performance for each motion across the different
textures and summarize the results in Table~\ref{tab:planar2}.
We observe that scene flow-based stereo video deblurring outperforms  
all other methods for all motions.
In particular, averaged over all textures, it outperforms kernels from
2D ground truth flow also in the case of the 2D `\emph{upward}' motion.
This is due to the fact that homographies are invertible and do not
suffer from the same rasterization artifacts as kernel generation
through forward warping with optical flow.
\input{tab_planar2}
\input{fig_planar2}
\section{Evaluation of the Structural Similarity Index}
In the main paper we report the peak-signal-to-noise-ratio (PSNR) between sharp reference images 
and deblurred images.
However, to be \eg consistent with human perception, also other evaluation measures for deblurred images are used.
In Table~\ref{table:occlusion2} we evaluate the Structural Similarity (SSIM) index~\cite{wang2004image} between deblurred images and sharp reference images.
The evaluation confirms the favorable results of visual inspection and the evaluation with the PSNR:
In all but the \emph{'apples'} scene, the use of 3D homographies increases SSIM in comparison to using any form of 2D displacement for blur kernel generation.
By suppressing locations with invalid image formation model, the influence of motion boundaries and erroneous scene flow computation can be compensated for and our full algorithm obtains consistently better results than deblurring based on 2D displacements.
\input{tab_occlusion2}



%% file: fig_pptKernels.tex
\begin{figure}[tbh]
\centering
%
%
\subfloat[3D motion]{\label{fig:pptKernel:3D}
\includegraphics[width = 0.33\columnwidth]{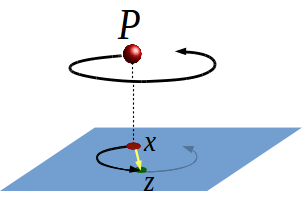} }%
\subfloat[Image plane projection: \newline Blur kernels from linear displacements]{\label{fig:pptKernel:2D}\includegraphics[width = 0.33\columnwidth]{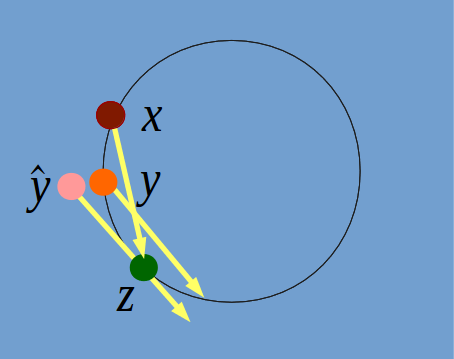} }%
\subfloat[Image plane projection: \newline Blur kernels from homographies]{\label{fig:pptKernel:hom}\includegraphics[width = 0.33\columnwidth]{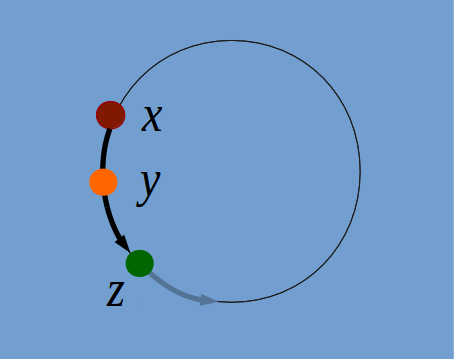} }%
\caption{Inducing blur matrices:
  \protect\subref{fig:pptKernel:3D}~Assume that 3D point $P$ moves
  with a constant rigid body motion in 3D, \eg with a yaw motion. The
  projection of this motion to the image plane (blue) is a circular
  trajectory. The corresponding 2D ground truth displacement (yellow),
  however, is a vector in the image plane that connects start and end
  point of the motion during a time interval.
  \protect\subref{fig:pptKernel:2D}~Using optical flow, the blur
  kernel at a location $z$ is approximated by identifying all pixels
  that, according to their spatially-variant displacement, pass
  through $z$ during the time interval.
  The image content at point $x$ is correctly identified as passing
  through $z$. The image content at point $y$ is not identified correctly as its
  2D displacement passes $z$ at a distance. 
  Instead, the image content at point $\hat{y}$ is erroneously
  identified as passing through $z$, even though $\hat{y}$ has a
  different distance to the rotation center than $z$.
  This results in the distorted kernels shown in Fig.~\ref{fig:kernel:rotFlow}.
  \protect\subref{fig:pptKernel:hom}~Assuming 3D points in the
  vicinity of $P$ to form a plane, we employ 3D homographies to generate
  blur kernels.
  The blur kernel at location $z$ is thus formed by the image points
  whose trajectory, according to the homography, passes through $z$
  during the time interval.
  Consequently, $y$ is correctly identified as passing through $z$,
  leading to the kernels shaped as circular arcs shown in Fig.~\ref{fig:kernel:rotHom}
}
\label{fig:pptKernels}
\end{figure}

%% file: fig_sceneFlow2.tex
\begin{figure}[t]
\centering
\includegraphics[width = 0.24\columnwidth]{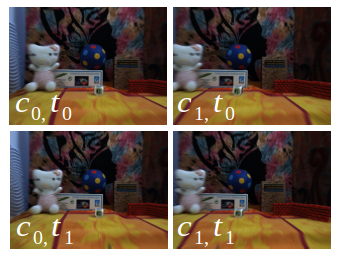} %
\includegraphics[width = 0.24\columnwidth]{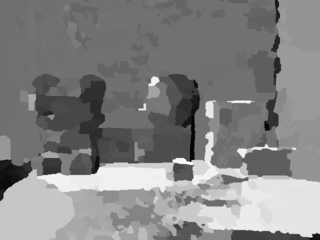} %
\includegraphics[width = 0.24\columnwidth]{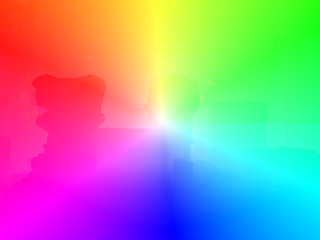} %
\includegraphics[width = 0.24\columnwidth]{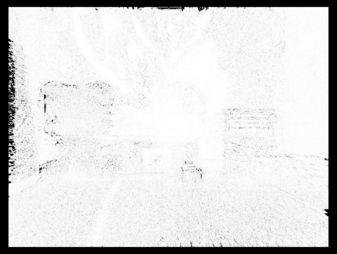} %
\\
\vspace{-0.35cm}
\subfloat[Input]{\label{fig:sceneFlow2:input}\includegraphics[width = 0.24\columnwidth]{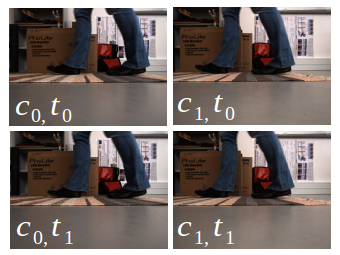} }%
\subfloat[Over-segmentation]{\label{fig:sceneFlow2:overseg}\includegraphics[width = 0.24\columnwidth]{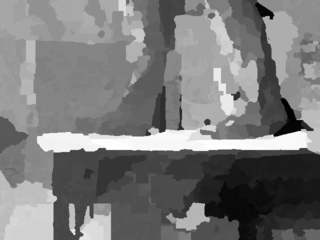} }%
\subfloat[2D projection of scene flow]{\label{fig:sceneFlow2:OF}\includegraphics[width = 0.24\columnwidth]{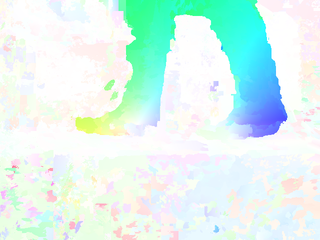} }%
\subfloat[Final occlusion weights]{\label{fig:sceneFlow2:weights}\includegraphics[width = 0.24\columnwidth]{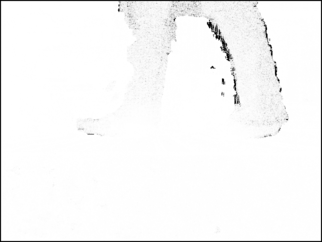} }%
\caption{Stereo video deblurring: For two consecutive frames of stereo
  video~\protect\subref{fig:sceneFlow2:input} we use the scene flow  
  approach of Vogel \etal~\cite{Vogel:2015:3SF} to compute an
  over-segmentation into planar patches with constant 3D rigid body
  motion~\protect\subref{fig:sceneFlow2:overseg}. For the real scenes shown, we obtain $1000-1500$ segments per frame-pair.
  For adjacent planes, the projection onto the image plane results in
  smoothly varying optical flow~\protect\subref{fig:sceneFlow2:OF}. At
  object boundaries, discontinuities in the flow coincide with segment boundaries. We use occlusion information from the scene flow to initialize object boundary weights. The weights are refined iteratively to downweight pixels that do not satisfy our image formation model~\protect\subref{fig:sceneFlow2:weights}
}
\label{fig:sceneFlow2}
\end{figure}

%% file: tab_planar2.tex
\begin{table}[bth]
\caption{Deblurring planar textures without motion discontinuities: 
  We report the peak signal-to-noise ratio (PSNR) of the deblurred
  reference frame, the average endpoint error (AEP) of the optical
  flow, and the average disparity error (ADE).
  All values are averaged over 8 different planar textures.  
  For all motions the use of scene flow increases deblurring accuracy
  compared to using 2D displacements. Best results (bold) are obtained with homography based deblurring
  } 
\label{tab:planar2}
\smallskip
\centering
\setlength{\tabcolsep}{8pt}
\begin{tabu}{@{}l*7{c}@{}}
\toprule
blur kernel & \multicolumn{1}{c}{ground truth} &
\multicolumn{2}{c}{initial optical} & \multicolumn{2}{c}{2D projection}  & \multicolumn{2}{c@{}}{3D}\\
source & \multicolumn{1}{c}{2D displacement} & \multicolumn{2}{c}{flow \cite{Vogel:2013:EDC}} & \multicolumn{2}{c}{of scene flow}  & \multicolumn{2}{c@{}}{homographies}
\\
\cmidrule(lr){2-2}
\cmidrule(lr){3-4}
\cmidrule(lr){5-6}
\cmidrule(lr){7-8}
& PSNR & AEP & PSNR & AEP & PSNR & ADE & PSNR\\
 \midrule
forward + roll  &32.32 &0.10 &32.64 &0.10 &32.73 &0.20 &{\bf 33.57}\\
forward         &34.22 &0.10 &34.60 &0.09 &34.83 &0.22 &{\bf 35.40}\\
upward          &26.94 &0.37 &26.75 &0.17 &26.88 &0.97 &{\bf 26.95}\\
forward + yaw   &29.40 &0.15 &29.43 &0.14 &29.43 &0.40 &{\bf 30.56}\\
yaw             &26.86 &3.01 &25.37 &0.47 &26.81 &0.28 &{\bf 27.32}\\
lateral + pitch&28.26 &0.15 &28.13 &0.20 &28.23 &0.02 &{\bf 29.18}\\
\bottomrule
\end{tabu}

\end{table}
\setlength{\tabcolsep}{1.4pt}

%% file: fig_planar2.tex
\begin{figure}[tbh]
\centering
\hspace{-0.2cm}
\includegraphics[width = 0.19\columnwidth]{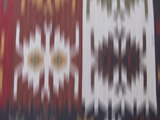} %
\includegraphics[width = 0.19\columnwidth]{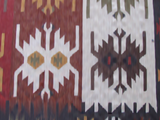} %
\includegraphics[width = 0.19\columnwidth]{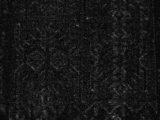} %
\includegraphics[width = 0.19\columnwidth]{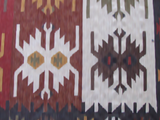} %
\includegraphics[width = 0.19\columnwidth]{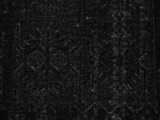} %
\\
\hspace{-0.2cm}
\includegraphics[width = 0.19\columnwidth]{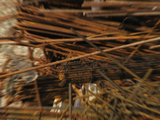} %
\includegraphics[width = 0.19\columnwidth]{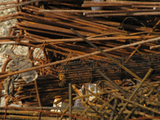} %
\includegraphics[width = 0.19\columnwidth]{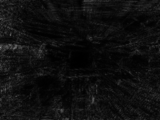} %
\includegraphics[width = 0.19\columnwidth]{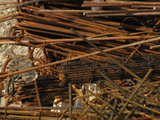} %
\includegraphics[width = 0.19\columnwidth]{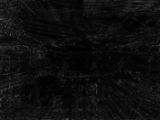} %
\\
\hspace{-0.2cm}
\includegraphics[width = 0.19\columnwidth]{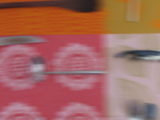} %
\includegraphics[width = 0.19\columnwidth]{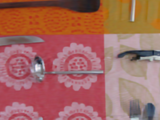} %
\includegraphics[width = 0.19\columnwidth]{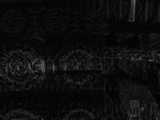} %
\includegraphics[width = 0.19\columnwidth]{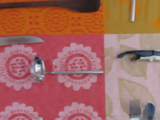} %
\includegraphics[width = 0.19\columnwidth]{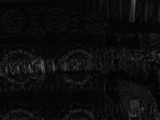} %
\\
\hspace{-0.2cm}
\includegraphics[width = 0.19\columnwidth]{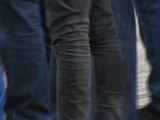} %
\includegraphics[width = 0.19\columnwidth]{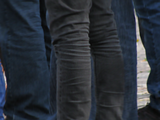} %
\includegraphics[width = 0.19\columnwidth]{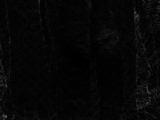} %
\includegraphics[width = 0.19\columnwidth]{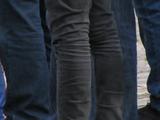} %
\includegraphics[width = 0.19\columnwidth]{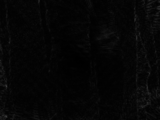} %
\\
\hspace{-0.2cm}
\includegraphics[width = 0.19\columnwidth]{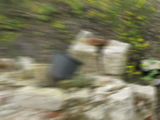} %
\includegraphics[width = 0.19\columnwidth]{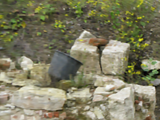} %
\includegraphics[width = 0.19\columnwidth]{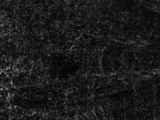} %
\includegraphics[width = 0.19\columnwidth]{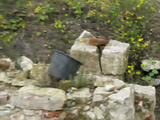} %
\includegraphics[width = 0.19\columnwidth]{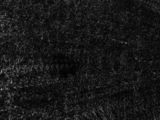} %
\\
\hspace{-0.2cm}
\includegraphics[width = 0.19\columnwidth]{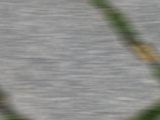} %
\includegraphics[width = 0.19\columnwidth]{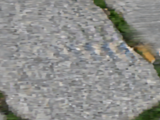} %
\includegraphics[width = 0.19\columnwidth]{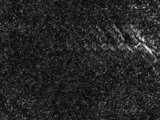} %
\includegraphics[width = 0.19\columnwidth]{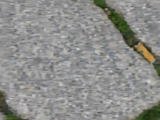} %
\includegraphics[width = 0.19\columnwidth]{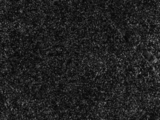} %
\\
\vspace{-0.35cm}
\subfloat[Input]{\label{fig:planar2:input}\includegraphics[width = 0.19\columnwidth]{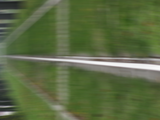} }%
\subfloat[Using kernels from \newline initial optical flow]{\label{fig:planar2:GT}\includegraphics[width = 0.19\columnwidth]{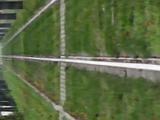} }%
\subfloat[Difference~(b) to\newline reference image]{\label{fig:planar2:GTdiff}\includegraphics[width = 0.19\columnwidth]{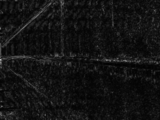} }%
\subfloat[Using kernels \newline from homographies]{\label{fig:planar2:3SF}\includegraphics[width = 0.19\columnwidth]{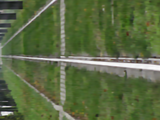} }%
\subfloat[Difference~(d) to\newline reference image]{\label{fig:planar2:3SFdiff}\includegraphics[width = 0.19\columnwidth]{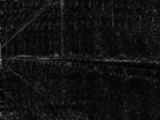} }%
\caption{Deblurring various planar textures: For a planar texture blurred with 3D rigid body
  motion~\protect\subref{fig:planar2:input}, deblurring with initial optical flow estimates~\protect\subref{fig:planar2:GT}
  yields differences~\protect\subref{fig:planar2:GTdiff} to the sharp reference image due to wrongly estimated flow and due to approximated kernels. These errors can be reduced by
  deblurring with our image formation model that uses scene flow to compute more robust motion and more accurate blur matrices~\protect\subref{fig:planar2:3SF},~\protect\subref{fig:planar2:3SFdiff}
}
\label{fig:planar2}
\end{figure}

%% file: tab_occlusion2.tex
\setlength{\tabcolsep}{6pt}
\begin{table}[hbt]
\caption{Deblurring with motion discontinuities:
While Tab.~\ref{table:occlusion} evaluates PSNR as error measure, here the Structural Similarity (SSIM) index~\cite{wang2004image} of deblurred synthetic scenes with motion discontinuities (top) and real scenes with the camera moving on a motorized rail (bottom) is evaluated.
Also for this evaluation measure, our homography-based stereo video deblurring with motion boundary weighting (full) clearly outperforms monocular video deblurring with optical-flow induced blur kernels}
\label{table:occlusion2}
\centering
\setlength{\tabcolsep}{9pt}
\begin{tabu}{@{}lccccccccc@{}}
\toprule
& \multicolumn{1}{c}{initial optical} & \multicolumn{1}{c}{2D projection} & \multicolumn{1}{c}{3D} & \multicolumn{1}{c}{ours}  & \multicolumn{1}{c@{}}{}
\\
& \multicolumn{1}{c}{flow \cite{Vogel:2013:EDC}} & \multicolumn{1}{c}{of scene flow} & \multicolumn{1}{c}{homographies} & \multicolumn{1}{c}{(full)}  & \multicolumn{1}{c@{}}{Kim and Lee \cite{Kim:2015:GVD}}\\
\midrule
apples           &0.881&  0.939& 0.928 &{\bf 0.950}& 0.928 \\
bunny            &0.770&  0.797& 0.826 &{\bf 0.834}& 0.720 \\
chair            &0.817&  0.853& 0.876 &{\bf 0.905}& 0.835 \\
squares          &0.709&  0.776& 0.810 &{\bf 0.869}& 0.796 \\
triplane         &0.818&  0.830& 0.866 &{\bf 0.869}& 0.693 \\
\hline
bottles          &0.955&  0.953& {\bf 0.972} &{\bf 0.972}& 0.942 \\
office           &0.958&  0.956& {\bf 0.968} &{\bf 0.968}& 0.935 \\
planar           &0.961&  0.961&       0.975 &{\bf 0.976}& 0.945 \\
toys             &0.953&  0.954& {\bf 0.967} &{\bf 0.967}& 0.931 \\
\bottomrule
\end{tabu}

\end{table}
\setlength{\tabcolsep}{1.4pt}

%% file: arxiv_deblurSceneFlow.bbl
\begin{thebibliography}{10}

\bibitem{Longuet:1981:CAR}
Longuet-Higgins, H.C.:
\newblock A computer algorithm for reconstructing a scene from two projections.
\newblock Nature \textbf{293} (September 1981)  133--135

\bibitem{Shotton:2013:EHP}
Shotton, J., Girshick, R., Fitzgibbon, A., Sharp, T., Cook, M., Finocchio, M.,
  Moore, R., Kohli, P., Criminisi, A., Kipman, A., Blake, A.:
\newblock Efficient human pose estimation from single depth images.
\newblock IEEE T. Pattern Anal. Mach. Intell. \textbf{35}(12) (December 2013)
  2821--2840

\bibitem{Franke:2000:RTS}
Franke, U., Joos, A.:
\newblock Real-time stereo vision for urban traffic scene understanding.
\newblock In: Intelligent Vehicles Symposium. (2000)  273--278

\bibitem{Geiger:2012:AWR}
Geiger, A., Lenz, P., Urtasun, R.:
\newblock Are we ready for autonomous driving? {T}he {KITTI} vision benchmark
  suite.
\newblock In: CVPR 2012.  3354--3361

\bibitem{Menze:2015:OSF}
Menze, M., Geiger, A.:
\newblock Object scene flow for autonomous vehicles.
\newblock In: CVPR 2015.  3061--3070

\bibitem{Vogel:2015:3SF}
Vogel, C., Schindler, K., Roth, S.:
\newblock {3D} scene flow estimation with a piecewise rigid scene model.
\newblock Int. J. Comput. Vision \textbf{115}(1) (October 2015)  1--28

\bibitem{Kim:2015:GVD}
Hyun~Kim, T., Mu~Lee, K.:
\newblock Generalized video deblurring for dynamic scenes.
\newblock In: CVPR 2015.  5426--5434

\bibitem{li2010generating}
Li, Y., Kang, S.B., Joshi, N., Seitz, S.M., Huttenlocher, D.P.:
\newblock Generating sharp panoramas from motion-blurred videos.
\newblock In: CVPR 2010.  2424--2431

\bibitem{yahyanejad2010removing}
Yahyanejad, S., Strom, J.:
\newblock Removing motion blur from barcode images.
\newblock In: CVPR 2010.  41--46

\bibitem{Fergus:2006:RCS}
Fergus, R., Singh, B., Hertzmann, A., Roweis, S.T., Freeman, W.T.:
\newblock Removing camera shake from a single photograph.
\newblock SIGGRAPH 2006  787--794

\bibitem{Cho:2009:FMD}
Cho, S., Lee, S.:
\newblock Fast motion deblurring.
\newblock ACM T. Graphics \textbf{28}(5) (December 2009)  145:1--145:8

\bibitem{Whyte:2010:NDS}
Whyte, O., Sivic, J., Zisserman, A., Ponce, J.:
\newblock Non-uniform deblurring for shaken images.
\newblock In: CVPR 2010.  491--498

\bibitem{Krishnan:2011:BDN}
Krishnan, D., Tay, T., Fergus, R.:
\newblock Blind deconvolution using a normalized sparsity measure.
\newblock In: CVPR 2011.  233--240

\bibitem{Xu:2013:ULS}
Xu, L., Zheng, S., Jia, J.:
\newblock Unnatural ${L}_0$ sparse representation for natural image deblurring.
\newblock In: CVPR 2013.  1107--1114

\bibitem{Michaeli:2014:BDI}
Michaeli, T., Irani, M.:
\newblock Blind deblurring using internal patch recurrence.
\newblock In: ECCV 2014. Volume~1.  171--184

\bibitem{Schelten:2014:LIB}
Schelten, K., Roth, S.:
\newblock Localized image blur removal through non-parametric kernel
  estimation.
\newblock In: ICPR 2014.  702--707

\bibitem{Couzinie-Devy:2013:LER}
Couzini\'{e}-Devy, F., Sun, J., Alahari, K., Ponce, J.:
\newblock Learning to estimate and remove non-uniform image blur.
\newblock In: CVPR 2013.  1075--1082

\bibitem{Wulff:2014:MBV}
Wulff, J., Black, M.J.:
\newblock Modeling blurred video with layers.
\newblock In: ECCV 2014. Volume~1.
\newblock  236--252

\bibitem{tai2011richardson}
Tai, Y.W., Tan, P., Brown, M.S.:
\newblock Richardson-lucy deblurring for scenes under a projective motion path.
\newblock IEEE T. Pattern Anal. Mach. Intell. \textbf{33}(8) (2011)  1603--1618

\bibitem{gupta2010single}
Gupta, A., Joshi, N., Zitnick, C.L., Cohen, M., Curless, B.:
\newblock Single image deblurring using motion density functions.
\newblock In: ECCV 2010. Volume~1.
\newblock  171--184

\bibitem{rajagopalan2014motion}
Rajagopalan, A., Chellappa, R.:
\newblock Motion Deblurring: Algorithms and Systems.
\newblock Cambridge University Press (2014)

\bibitem{Levin:2006:BMD}
Levin, A.:
\newblock Blind motion deblurring using image statistics.
\newblock In: NIPS*2006.  841--848

\bibitem{Chakrabarti:2010:ASB}
Chakrabarti, A., Zickler, T., Freeman, W.T.:
\newblock Analyzing spatially-varying blur.
\newblock In: CVPR 2010.  2512--2519

\bibitem{Kim:2013:DSD}
Kim, T.H., Ahn, B., Lee, K.M.:
\newblock Dynamic scene deblurring.
\newblock In: ICCV 2013.  3160--3167

\bibitem{Sun:2015:LCN}
Sun, J., Cao, W., Xu, Z., Ponce, J.:
\newblock Learning a convolutional neural network for non-uniform motion blur
  removal.
\newblock In: CVPR 2015.  769--777

\bibitem{Kim:2014:SFD}
Kim, T.H., Lee, K.M.:
\newblock Segmentation-free dynamic scene deblurring.
\newblock In: CVPR 2014.  2766--2773

\bibitem{Xu:2012:DAM}
Xu, L., Jia, J.:
\newblock Depth-aware motion deblurring.
\newblock In: ICCP. (2012)  1--8

\bibitem{lee2013dense}
Lee, H., Lee, K.:
\newblock Dense 3d reconstruction from severely blurred images using a single
  moving camera.
\newblock In: CVPR 2013.  273--280

\bibitem{arun2015multi}
Arun, M., Rajagopalan, A., Seetharaman, G.:
\newblock Multi-shot deblurring for 3d scenes.
\newblock In: CVPR 2015.  19--27

\bibitem{hu2014joint}
Hu, Z., Xu, L., Yang, M.H.:
\newblock Joint depth estimation and camera shake removal from single blurry
  image.
\newblock In: CVPR 2014.  2893--2900

\bibitem{cho2007removing}
Cho, S., Matsushita, Y., Lee, S.:
\newblock Removing non-uniform motion blur from images.
\newblock In: ICCV 2007.  1--8

\bibitem{he2010motion}
He, X., Luo, T., Yuk, S., Chow, K., Wong, K.Y., Chung, R.:
\newblock Motion estimation method for blurred videos and application of
  deblurring with spatially varying blur kernels.
\newblock In: ICCIT. (2010)  355--359

\bibitem{deng2012video}
Deng, X., Shen, Y., Song, M., Tao, D., Bu, J., Chen, C.:
\newblock Video-based non-uniform object motion blur estimation and deblurring.
\newblock Neurocomputing \textbf{86} (2012)  170--178

\bibitem{yamaguchi2010video}
Yamaguchi, T., Fukuda, H., Furukawa, R., Kawasaki, H., Sturm, P.:
\newblock Video deblurring and super-resolution technique for multiple moving
  objects.
\newblock In: Computer Vision--ACCV 2010.
\newblock Springer (2010)  127--140

\bibitem{vedula1999three}
Vedula, S., Baker, S., Rander, P., Collins, R., Kanade, T.:
\newblock Three-dimensional scene flow.
\newblock In: ICCV 1999. Volume~2.  722--729

\bibitem{quiroga2013local}
Quiroga, J., Devernay, F., Crowley, J.:
\newblock Local/global scene flow estimation.
\newblock In: ICIP. (2013)  3850--3854

\bibitem{wedel2011stereo}
Wedel, A., Cremers, D.:
\newblock Stereo scene flow for 3D motion analysis.
\newblock Springer Science \& Business Media (2011)

\bibitem{Tai:2010:Correction}
Tai, Y.W., Du, H., Brown, M.S., Lin, S.:
\newblock Correction of spatially varying image and video motion blur using a
  hybrid camera.
\newblock IEEE T. Pattern Anal. Mach. Intell. \textbf{32}(6) (2010)  1012--1028

\bibitem{chen2008robust}
Chen, J., Yuan, L., Tang, C.K., Quan, L.:
\newblock Robust dual motion deblurring.
\newblock In: CVPR 2008.  1--8

\bibitem{cho2012video}
Cho, S., Wang, J., Lee, S.:
\newblock Video deblurring for hand-held cameras using patch-based synthesis.
\newblock ACM T. Graphics \textbf{31}(4) (2012)  64:1--64:9

\bibitem{joshi2010image}
Joshi, N., Kang, S.B., Zitnick, C.L., Szeliski, R.:
\newblock Image deblurring using inertial measurement sensors.
\newblock ACM T. Graphics \textbf{29}(4) (2010)  30:1--30:9

\bibitem{mei2008modeling}
Mei, C., Reid, I.:
\newblock Modeling and generating complex motion blur for real-time tracking.
\newblock In: CVPR 2008.  1--8

\bibitem{murray1994mathematical}
Murray, R.M., Li, Z., Sastry, S.S.:
\newblock A mathematical introduction to robotic manipulation.
\newblock CRC press (1994)

\bibitem{portz2012optical}
Portz, T., Zhang, L., Jiang, H.:
\newblock Optical flow in the presence of spatially-varying motion blur.
\newblock In: CVPR 2012.  1752--1759

\bibitem{Levin:2007:UAS}
Levin, A., Weiss, Y.:
\newblock User assisted separation of reflections from a single image using a
  sparsity prior.
\newblock IEEE T. Pattern Anal. Mach. Intell. \textbf{29}(9) (September 2007)
  1647--1654

\bibitem{Vogel:2013:EDC}
Vogel, C., Schindler, K., Roth, S.:
\newblock An evaluation of data costs for optical flow.
\newblock In: Pattern Recognition (GCPR) 2013.  343--353

\bibitem{wang2004image}
Wang, Z., Bovik, A.C., Sheikh, H.R., Simoncelli, E.P.:
\newblock Image quality assessment: {F}rom error visibility to structural
  similarity.
\newblock IEEE T. Image Process. \textbf{13}(4) (2004)  600--612

\end{thebibliography}
